\documentclass[preprint, 12pt]{elsarticle}

\usepackage{amssymb}

\usepackage{amsmath}
\usepackage{multirow}

\graphicspath{{figures}}

\journal{Knowledge-Based Systems}

\begin{document}

\begin{frontmatter}



\title{Image Projective Transformation Rectification with Synthetic Data for Smartphone-captured Chest X-ray Photos Classification }


\author[MPU]{Chak Fong Chong}
\ead{chakfong.chong@mpu.edu.mo}
\author[MPU]{Yapeng Wang}
\ead{yapengwang@mpu.edu.mo}
\author[MPU]{Benjamin Ng}
\ead{bng@mpu.edu.mo}
\author[MPU]{Wuman Luo}
\ead{wumanluo@mpu.edu.mo}
\author[MPU]{Xu Yang\corref{cor}}
\ead{xuyang@mpu.edu.mo}

\cortext[cor]{Corresponding author}


\address[MPU]{Macao Polytechnic University, Macao SAR, China}

\begin{abstract}

Classification on smartphone-captured chest X-ray (CXR) photos to detect pathologies is challenging due to the projective transformation caused by the non-ideal camera position. Recently, various rectification methods have been proposed for different photo rectification tasks such as document photos, license plate photos, etc. Unfortunately, we found that none of them is suitable for CXR photos, due to their specific transformation type, image appearance, annotation type, etc. In this paper, we propose an innovative deep learning-based \textbf{P}rojective \textbf{T}ransformation \textbf{R}ectification \textbf{N}etwork (PTRN) to automatically rectify CXR photos by predicting the projective transformation matrix. To the best of our knowledge, it is the first work to predict the projective transformation matrix as the learning goal for photo rectification. Additionally, to avoid the expensive collection of natural data, synthetic CXR photos are generated under the consideration of natural perturbations, extra screens, etc. We evaluate the proposed Method in the CheXphoto smartphone-captured CXR photos classification competition hosted by the Stanford University Machine Learning Group, our approach won \textbf{first place} with a huge performance improvement (ours 0.850, second-best 0.762, in AUC). A deeper study demonstrates that the use of PTRN successfully achieves the classification performance on the spatially transformed CXR photos to the same level as on the high-quality digital CXR images, indicating PTRN can eliminate all negative impacts of projective transformation on the CXR photos.

\end{abstract}




\end{frontmatter}


\section{Introduction} \label{sec:Introduction}

Chest X-ray (CXR), also named chest radiograph, is one of the most ubiquitous medical imaging techniques for chest disease diagnosis. However, due to the lack of resources, many CXRs cannot be formally reviewed by radiologists on time \cite{care2018radiology, royal2015unreported}. For instance, in only one hospital in the United Kingdom, over 23,000 CXRs were not formally reviewed on time in the past 12 months \cite{baltruschat2019comparison}. Such serious situations provide a strong motivation to develop computer-aid diagnosis (CAD) systems for CXR automatic interpretation.

Convolutional neural networks (CNN) have been used to perform CXR multi-label classification tasks to detect pathologies. These CNN classification models are usually trained on datasets of digital CXRs \cite{wang2017chestx,irvin2019chexpert,johnson2019mimic} produced by modern computer-based digital X-ray systems \cite{rajpurkar2017chexnet,pham2021interpreting,yuan2021large,kuo2021recalibration,guan2021discriminative}. The classification performances are promising and even approach radiologist-level \cite{rajpurkar2017chexnet,yuan2021large}.

Unfortunately, in many developing countries and regions, traditional X-ray film systems are still widely used \cite{schwartz2014accuracy,andronikou2011paediatric}. CXR films are hard copies, so cannot be interpreted directly by computer-based models. To address this problem, a quick and convenient solution is to use smartphones to capture photos of CXR films (\textit{CXR film photos}), then a CXR classification model interprets the captured CXR photos \cite{kuo2021recalibration,phillips2020chexphoto, rajpurkar2021chexternal, chong2021gan}. Benefitting from the low prices and popularity of smartphones, it is a cheap and effective solution for countries and regions which have very limited medical resources. Moreover, the use of smartphones can facilitate both real-time and store-and-forward medical consultation to provide remote medical care and remote diagnosis \cite{liu2020application, karako2020realizing}. Especially, the sudden outbreak of COVID-19 in 2020 has promoted the practical applications of telemedicine and zero-contact diagnosis. Furthermore, it can help patients who have privacy concerns that not willing to go to hospitals for radiology reports.

Additionally, to demonstrate the generalization of our proposed method on CXR photos captured in various unconstrainted scenarios, we also include smartphone-captured photos of digital CXR images displayed on the monitor screens (\textit{CXR monitor photos}) to evaluate our proposed method in this paper.

\begin{figure}
    \centering
    \includegraphics[width=0.5\textwidth]{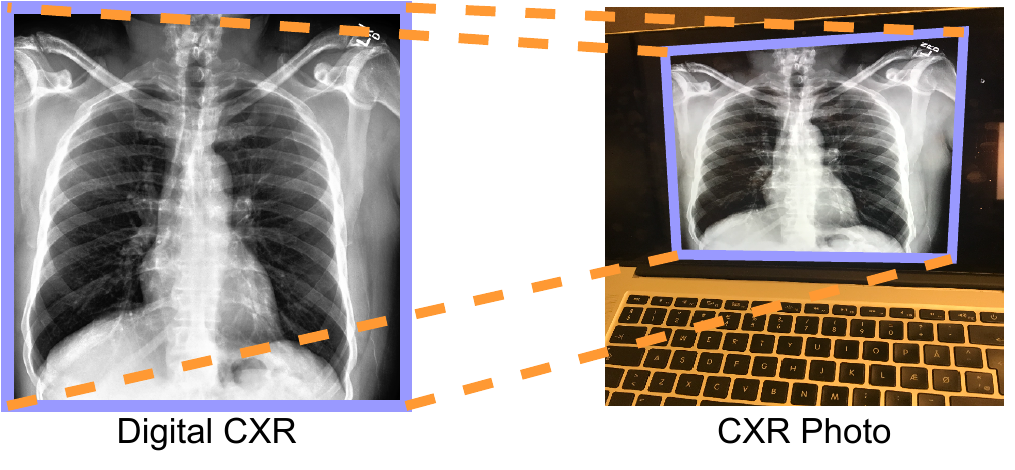}
    \caption{The CXR in a photo are warped by projective transformation due to non-ideal camera position.}
    \label{fig:projective_transformation}
\end{figure}

The classification on smartphone-captured CXR photos is a very challenging task because CXR photos have very different appearances from digital CXRs, which leads to a significant performance drop. In a CXR photo, (1) the CXR is warped by projective transformation due to non-ideal camera position \cite{liang2005camera}, as shown in Figure \ref{fig:projective_transformation}; (2) Natural perturbations such as environmental illuminations, camera out-focus, image noises, etc., exist in photos. Moreover, CXR photos are captured in various unconstrained scenarios, thus they have very different projective transformations and natural perturbations. Therefore, the classification performance drops significantly when a CNN model is trained on high-quality digital CXR images and then evaluated on CXR photos \cite{kuo2021recalibration,chong2021gan,rajpurkar2021chexternal}. In the paper of \cite{chong2021gan}, Chong et al. demonstrated that the projective transformation of CXR photos is the major reason for the significant classification performance drop.

Therefore, one obvious solution to improve classification performance on CXR photos is to automatically rectify the projective transformation of CXR photos. Chong et al. \cite{chong2021gan} proposed a method named generative adversarial network-based spatial transformation adversarial method (GAN-STAM)  to automatically rectify CXR photos. However, the results showed that GAN-STAM cannot properly rectify CXR photos, probably due to the following reasons: the lack of training data, inappropriate rectification mechanism (affine transformation), instability of GAN training, etc.

Recently, various rectification methods have been proposed for different photo rectification tasks such as document photos \cite{xue2022fourier,jiang2022revisiting,bandyopadhyay2022rectinet,ma2018docunet,das2019dewarpnet,feng2021docscanner,quan2022recovering}, license plate photos \cite{silva2021flexible,bjorklund2019robust,silva2018license, xu20212d, xu2021eilpr}, etc. Unfortunately, we found that none of them is suitable for smartphone-captured CXR photos, due to their specific transformation type, image appearance, annotation type, etc. Normally, an existing method is particularly designed for one specific rectification task only. For example, in document photos unwarping, the documents in photos are folded and curved, which are not suitable for smartphone-captured CXR photos.

In this paper, we propose an innovative method to rectify the projective transformation of CXR photos caused by non-ideal camera position, named \textbf{P}rojective \textbf{T}ransformation \textbf{R}ectification \textbf{N}etwork (PTRN). To the best of our knowledge, it is the first work to predict the projective transformation matrix as the learning goal for photo rectification. Additionally, synthetic CXR photos with transformation matrices as the ground truth annotations are generated for training PTRN. Moreover, the designs of both PTRN and the synthetic data generation framework are general to any camera-captured photos such as CCTV-captured license plate photos.

The primary contributions of this work are summarized:

\begin{enumerate}
    \item We propose an innovative method named PTRN to rectify the projective transformation caused by non-ideal camera positions. The transformation matrix is set as the learning goal of PTRN for photo rectification. Additionally, the Intersection over Union (IoU), normally used to quantify the percent overlap of image segmentation, is first proposed as the performance metric to quantitively evaluate the rectification performance, to the best of our knowledge.
    \item We also propose the innovative synthetic CXR photos generation framework, which is designed under the consideration of the appearances of natural CXR photos including screen, background, and natural perturbations, to avoid the collection and annotation of expensive natural data. The framework produces projective transformation matrices as the ground truth annotations, instead of bounding boxes which are common in synthetic data generation frameworks for other object detection tasks. The proposed framework is fast yet efficient, as it only uses general image processing methods.
    \item Specifically, we design a CXR photos classification pipeline in three steps: (1) PTRN predicts the projective transformation matrix of a CXR photo; (2) The photo is rectified using the predicted matrix; (3) A classifier trained on high-quality digital CXRs evaluates the rectified photo to produce the final result.
    \item We train PTRN on synthetic data generated using the CheXpert digital CXR dataset \cite{irvin2019chexpert} and the MS-COCO dataset \cite{lin2014microsoft}. We also train a classifier on the CheXpert dataset to build up the CXR photos classification pipeline. Then, the pipeline performance is evaluated on the CheXphoto smartphone-captured CXR photos classification dataset \cite{phillips2020chexphoto}. The pipeline achieves AUC 0.880/0.802 on the CXR monitor photos / CXR film photos, respectively, which is much higher than the performance without using PTRN to rectify photos (AUC 0.710/0.599). 
    \item The proposed PTRN with the classification pipeline is also evaluated on the CheXphoto competition hosted by Stanford and Vinbrain. Our pipeline achieves \textbf{first place} in the competition with AUC 0.850 which is much higher than the second-best approach (AUC 0.762) on the competition leaderboard. A deeper study demonstrates that the use of PTRN successfully achieves the classification performance on the spatially transformed CXR photos to the same level as on the high-quality digital CXR images, indicating PTRN can eliminate all negative impacts of projective transformation on the CXR photos.
\end{enumerate}

The rest of the paper is organized as follows. In Section \ref{sec:Related Work}, work related to smartphone-captured CXR photos classification, image rectification, and deep learning with synthetic data is introduced. Section \ref{sec: Methods} introduces PTRN and the synthetic data framework. Section \ref{sec:Experimental Results and Discussion} shows the experiment results and discussion. Section \ref{sec:Conclusion} is the conclusion, limitations, and future work.

\section{Related Work} \label{sec:Related Work}
\subsection{Smartphone-captured CXR Photos Classification}
Classification performance drop when current CXR classification models are evaluated on smartphone-captured CXR photos \cite{kuo2021recalibration,chong2021gan,rajpurkar2017chexnet}. It is because CXR photos have very different appearances from digital CXRs that the models are trained on, as introduced in Section \ref{sec:Introduction}.

Several methods have been proposed to improve classification performance. One first attempt is Le et al. \cite{le2020interpretation} proposed a method to crop the CXRs in photos using YOLO-v3 \cite{redmon2018yolov3}, but the classification performance is unacceptably low (AUC 0.684). Kuo et al. \cite{kuo2021recalibration} developed a recalibration method that uses data augmentation methods to add additional noises into digital CXRs for training. The method achieves AUC 0.835 on CXR photos. However, this method does not tackle the projective transformation problem. Chong et al. \cite{chong2021gan} found that once the projective transformation of CXR photos is perfectly rectified before classification, the classification performance significantly improves from AUC 0.801 to AUC 0.887. Therefore, they proposed GAN-STAM to automatically rectify the projective transformation. However, the classification performance attains AUC 0.865 only, which is still far away from the performance on perfectly rectified photos. It indicates GAN-STAM cannot properly rectify the projective transformation, probably due to these reasons: (1) GAN-STAM uses an affine transformation to rectify photos, but the CXRs in photos are warped by the projective transformation that is more complex than the affine transformation \cite{solomon2011fundamentals}. (2) the lack of data and annotations. (3) the training of GAN is unstable \cite{creswell2018generative,wang2017generative}; and (4) no quantitative evaluation metric to measure the rectification performance.

\subsection{Camera-captured Photos Rectification}

\textbf{Traditional model-based methods.}
Model-based methods have been proposed for document photos \cite{fang2011distortion,liang2008geometric,takezawa2017robust}, QR code photos \cite{chang2007general, tribak2017qr}, etc. These methods rely on image appearances. For document photos, horizontal sentences printed in the document are utilized by using image processing techniques like Radon transformation \cite{deans2007radon, takezawa2017robust} to calculate the distortion of the document such as the vanishing points and the level of curving. For a QR code photo, the QR code features (finder patterns and alignment patterns) are utilized to locate the QR codes for rectification. Overall, a traditional method is usually designed for rectifying a specific target. Additionally, these methods rely on low-level image patterns extracted by traditional digital image processing methods, so that unsuitable for photos captured in various unconstraint scenarios.

\textbf{Deep learning-based methods.}
Deep learning has been widely used for such rectification tasks since deep learning models can learn from a large amount of data. Therefore, it is more robust to camera-captured natural photos and reduces the manual design of algorithms. Various rectification methods have been proposed for different photo rectification tasks in the literature such as document photos, license plate photos, etc. Unfortunately, we found that none of them is suitable for CXR photos. It is because a task has its distinctive properties such as the transformation type, image appearance, annotation type, etc. Hence, a rectification method is particularly designed for a specific task and not compatible with other tasks. For document photo unwarping, the documents are folded and curved \cite{xue2022fourier,jiang2022revisiting,bandyopadhyay2022rectinet,ma2018docunet,das2019dewarpnet,feng2021docscanner,quan2022recovering}, which is unsuitable for CXR photos. In license plate recognition, several methods train models to crop/rectify the detected license plates in photos using affine transformation \cite{silva2021flexible,bjorklund2019robust,silva2018license}, but CXR photos are projective transformation which is more complex than affine transformation \cite{solomon2011fundamentals}. Some methods \cite{xu20212d, xu2021eilpr} utilized the spatial transformer network (STN) \cite{jaderberg2015spatial} (a learnable neural network module that can perform spatial transformations to feature maps or images) to reduce the problem of transformation. In these methods, A STN-like module is inserted into the license plate recognition network to reduce the transformation problem for better recognition accuracy. The whole recognition network is trained in an end-to-end manner. Therefore, the STN-like module aims to select the region in the whole image that the recognition network is interested in (attention), instead of precisely rectifying the license plate. Besides, some methods were proposed for other rectification tasks like barcode photos \cite{xiao20191d} and text photos \cite{zhan2019esir, xue2022detection}. The method for barcodes relies on the vertical lines of barcodes, while the transformation types of texts include line curvature which is very different from photos of CXR.

Overall, to the best of our knowledge, existing rectification methods are not compatible with the projective transformation of photos like CXR photos. Therefore, we propose PTRN which is the first method that predicts the projective transformation matrices to rectify the projective transformation of CXR photos caused by the non-ideal camera position.

\subsection{Synthetic Data for Photo Rectification and Object Detection}

Synthetic data is one popular method to address the problem of the lack of data for training deep learning models, as it avoids expensive natural data collection. Synthetic data has also been used for object detection \cite{gupta2016synthetic, dwibedi2017cut, zhan2018verisimilar} and camera-captured photo rectification tasks \cite{ma2018docunet,das2019dewarpnet,bjorklund2019robust}.

A common strategy for generating a synthetic image is a composition of a foreground image (i.e., a transformed image) and a background image, plus data augmentation to simulate natural perturbations \cite{gupta2016synthetic, dwibedi2017cut, zhan2018verisimilar,ma2018docunet,das2019dewarpnet,bjorklund2019robust}. However, the detailed generation steps are quite different across tasks, since a task has its distinctive properties such as natural appearance and annotation type. In text detection \cite{gupta2016synthetic}, a text usually appears in well-defined regions like a sign or a flat wall. Therefore, the generation step includes additional geometric estimation and segmentation to ensure the texts are properly placed in the correct positions. The annotation type of object detection is bounding box, while the annotation type of document photo unwarping is 3D coordinate maps \cite{das2019dewarpnet,ma2018docunet,feng2021docscanner}. Hence, the synthetic data generation steps are different.

Our proposed framework for synthetic data generation has a couple of novelties in comparison to previous synthetic data frameworks. We propose the first CXR photos synthetic framework. The design of the framework is under consideration of the natural appearances of CXR photos. Furthermore, our framework is the first that uses the projective transformation matrix as the ground truth annotation to represent the transformation caused by the non-ideal camera position. Unlike object detection tasks that use bounding boxes that only locate the object, the projective transformation matrix can be used to precisely rectify the photo. Which is innovative and effective.

\section{Methods} \label{sec: Methods}

We propose Projective Transformation Rectification Network (PTRN) for rectifying the projective transformation of CXR photos caused by non-ideal camera position. PTRN is trained in an end-to-end manner on synthetic CXR photos training samples. The designs of both PTRN and the synthetic data generation framework are general such that PTRN can also be used for rectifying other camera-captured photos of images (e.g., CCTV-captured license plate photos), detailed in follows.

\subsection{Problem Formulation}

This section formulates the problem of projective transformation rectification for CXR photos. The designs of PTRN and the synthetic data framework are based on the formulation.

\begin{figure}
    \centering
    \includegraphics[width=0.5\textwidth]{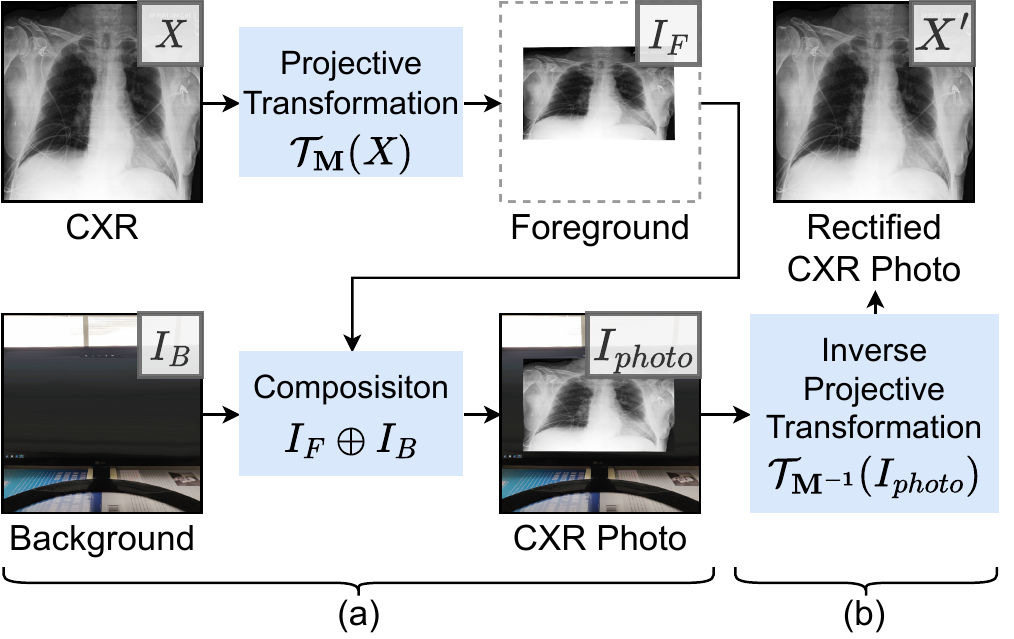}
    \caption{Formulation of projective transformation rectification for CXR photos.}
    \label{fig:formulation}
\end{figure}

A CXR photo $I_{photo}$ can be considered as a composition of a transformed CXR $I_F$ and a background image $I_B$, as in Figure \ref{fig:formulation} (a). We borrow a notation $\oplus$ from \cite{lin2018st} to represent the composition of two images. $I_{photo}$ can be formulated as:
\begin{equation}\label{eqn:1}
    I_{photo}=I_F \oplus I_B =\mathcal{T}_\mathbf{M}(X)\oplus I_B
\end{equation}
where $I_F$ is a CXR $X$ warped by projective transformation with a 3-by-3 transform matrix $\mathbf{M}$: $I_F=\mathcal{T}_\mathbf{M}(X)$. A matrix $\mathbf{M}$ has 8 parameters $\theta_i$ that represent a projective transformation: $\mathbf{M}=
\begin{bmatrix}
    \theta_1 & \theta_2 & \theta_3 \\
    \theta_4 & \theta_5 & \theta_6 \\
    \theta_7 & \theta_8 & 1 
\end{bmatrix}
$
. In this work, all projective transformations are applied in the homogeneous coordinates and the coordinates of the 4 vertices of each image are assigned to be $P_0=\begin{Bmatrix}
\begin{pmatrix} -1 \\ -1 \\1 \end{pmatrix},
\begin{pmatrix} 1 \\ -1 \\1 \end{pmatrix},
\begin{pmatrix} 1 \\ 1 \\1 \end{pmatrix},
\begin{pmatrix} -1 \\ 1 \\1 \end{pmatrix}
\end{Bmatrix}
$.

A CXR photo $I_{photo}$ can be rectified using inverse projective transformation if $\mathbf{M}$ is known (Figure \ref{fig:formulation} (b)):
\begin{equation}\label{eqn:2}
    X'=\mathcal{T}_{\mathbf{M}^{-1}}(I_{photo})
\end{equation}
where $X'$ denotes the rectified CXR photos.

The design of PTRN is based on Equation \ref{eqn:2}: A CXR photo $I_{photo}$ can be rectified by predicting $\mathbf{M}$. Therefore, the input of PTRN is a CXR photo $I_{photo}$ and the output is the predicted matrix $\hat{\mathbf{M}}$. Based on this design, a training sample for PTRN should consist of a CXR photo $I_{photo}$ and the matrix $\mathbf{M}$. To synthetically generate training samples, a generation framework is designed based on Equation \ref{eqn:1}: A training sample $(I_{photo},\mathbf{M})$ can be synthesized using a CXR $X$, a background image $I_B$ and $\mathbf{M}$. The designs of PTRN and the synthetic data generation framework are detailed in the following sections. Besides, we note this formulation is general to other camera-captured photos of images, not CXRs only, since the CXR appearance is not considered in the formulation. E.g., the formulation is also suitable for CCTV-captured license plate photos.

\subsection{Projective Transformation Rectification Network}

PTRN is a deep neural network that receives a CXR photo $I_{photo}$ and predicts the matrix $\mathbf{M}$ for rectification. PTRN is particularly designed not to rely on any appearances of CXRs. Hence, PTRN is also suitable for rectifying projective transformation of camera-captured photos of different image types.

\subsubsection{Architecture}

\begin{figure}
    \centering
    \includegraphics[width=0.5\textwidth]{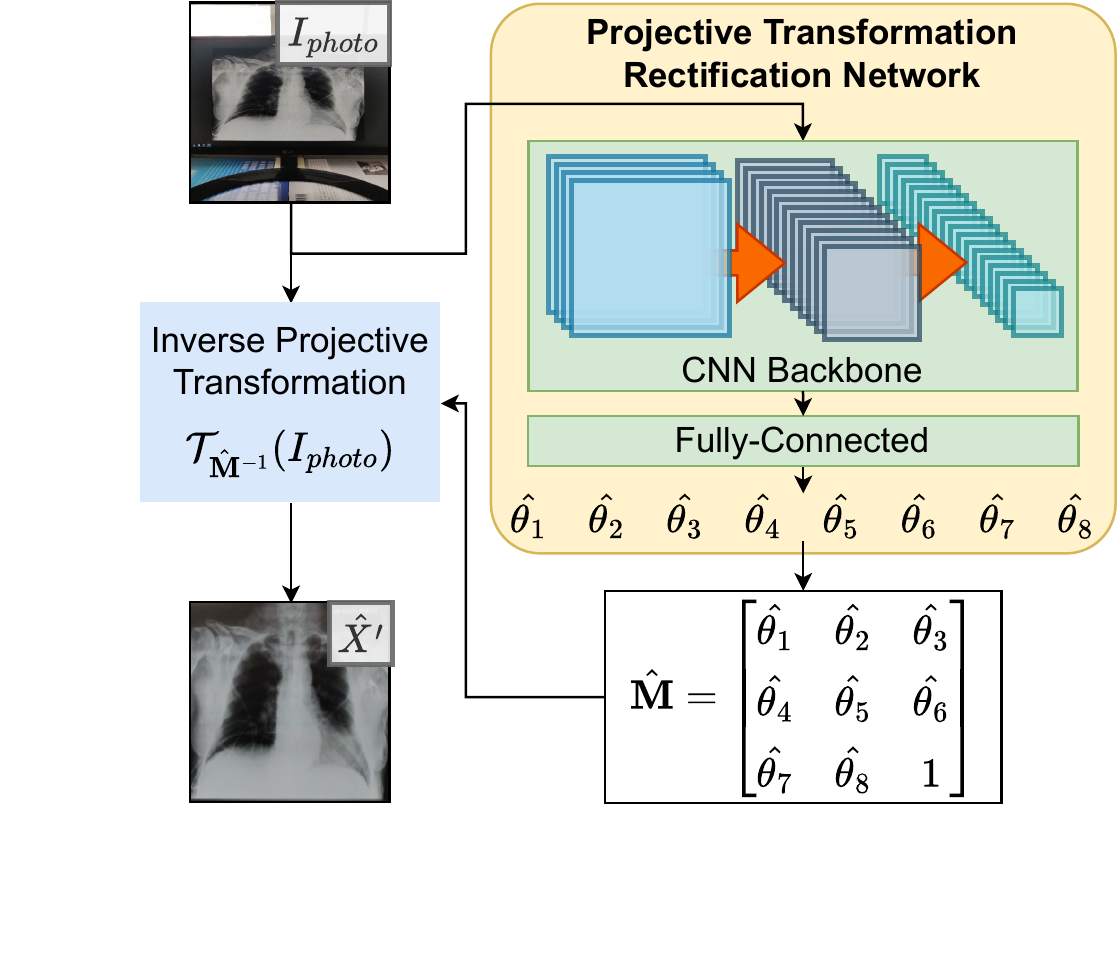}
    \caption{The architecture of PTRN and the steps of rectifying a CXR photo using PTRN.}
    \label{fig:PTRN_architecture}
\end{figure}

As shown in Figure \ref{fig:PTRN_architecture} (right). The architecture of PTRN consists of two components: (1) a CNN backbone (e.g., ResNet-50 \cite{he2016deep}, DenseNet-121 \cite{huang2017densely}), followed by (2) a fully-connected layer to regresses the 8 parameters $\theta_i$ of a predicted matrix $\hat{\mathbf{M}}$.

Rectifying a CXR photo $I_{photo}$ using PTRN has two steps, as shown in Figure \ref{fig:PTRN_architecture}: (1) PTRN predicts $\hat{\mathbf{M}}$ ; (2) Apply the inverse projective transformation with $\hat{\mathbf{M}}$ (Equation \ref{eqn:2}) to the CXR photo $I_{photo}$ .

\subsubsection{Training}

PTRN is trained on synthetic training samples $\{(I_{photo}^{(i)}, \mathbf{M}^{(i)})\}$ in an end-to-end manner. As PTRN predicts the parameters by regression, mean squared error (MSE) loss is used to calculate the loss between the model prediction and the ground truth. MSE loss is defined as $L=\frac{1}{N} \sum_{i=1}^N\left(y_i-\hat{y}_i\right)^2$ where $y$ denotes the 8 parameters of the ground truth $\mathbf{M}$: $y=\left[\theta_1 \dots \theta_8 \right]$ and $\hat{y}$ denotes the predicted 8  parameters $\hat{y}=[\hat{\theta_1} \dots \hat{\theta_8} ]$.

\subsubsection{CXR Region Prediction}

A matrix $\mathbf{M}$ can be converted to a quadrilateral that represents a CXR region. The 4 vertices $\begin{pmatrix}
    x_i \\ y_i \\ 1
\end{pmatrix}$ of the quadrilateral represented by $\mathbf{M}$ are:
\begin{equation} \label{eqn:3}
    \left(\begin{array}{c}
        w_i x_i \\
        w_i y_i \\
        w_i
        \end{array}\right)=\mathbf{M}\left(\begin{array}{c}
        u_i \\
        v_i \\
        1
        \end{array}\right)
\end{equation}
where $\begin{pmatrix}
    u_i \\ v_i \\ 1
\end{pmatrix} \in P_0$.

The quadrilateral can also be converted back to $\mathbf{M}$ using Equation \ref{eqn:3}. This relation is used in the evaluation of the rectification performance of PTRN. For instance, a predicted matrix $\hat{\mathbf{M}}$ can be converted to a predicted CXR region.

\subsubsection{Evaluation and Rectification Performance Metric}

PTRN is evaluated on natural test samples $\left\{\left(I_{\text {photo }}^{(i)}, P^{(i)}\right)\right\}$. Each sample consists of a natural CXR photo $I_{photo}$ and the 4 vertices $\left(\begin{array}{c}
    x_i \\
    y_i \\
    1
    \end{array}\right) \in P$ 
that are manually annotated to represent the ground-truth CXR region. The ground-truth label is the 4 vertices instead of the matrix $\mathbf{M}$ since marking the 4 vertices is the simplest way to annotate the ground truth in practice.

\begin{figure}
    \centering
    \includegraphics[width=0.5\textwidth]{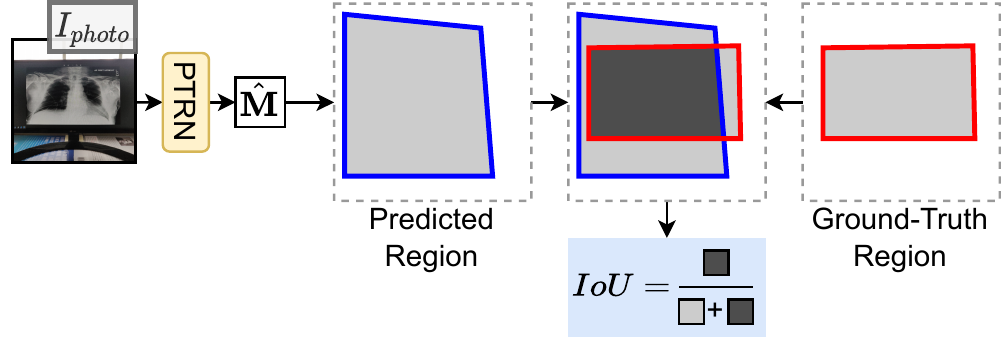}
    \caption{The rectification performance of PTRN is evaluated by the IoU between predicted regions and ground truth regions.}
    \label{fig:IoU}
\end{figure}

We also propose using IoU between the ground truth CXR region and the predicted CXR region for each sample to measure the rectification performance, as the predicted region should be as overlapped as the ground truth region. The predicted region is calculated from $\hat{\mathbf{M}}$ using Equation \ref{eqn:3} and the ground truth region is given in the test sample. The IoU calculation of a sample is demonstrated in Figure \ref{fig:IoU}.

\subsection{Training Data Synthesis}

PTRN requires a large amount of training data. Unfortunately, there are no suitable datasets. The collection of natural data samples is also expensive. Therefore, we propose a framework for the generation of synthetic training samples. To enable the generality of PTRN, the proposed framework is designed not to rely on the appearances of CXRs. Hence, the framework can also generate synthetic camera-captured photos of different image types with very few modifications. So that PTRN can be trained to rectify camera-captured photos of different images.

\begin{figure}
    \centering
    \includegraphics[width=0.5\textwidth]{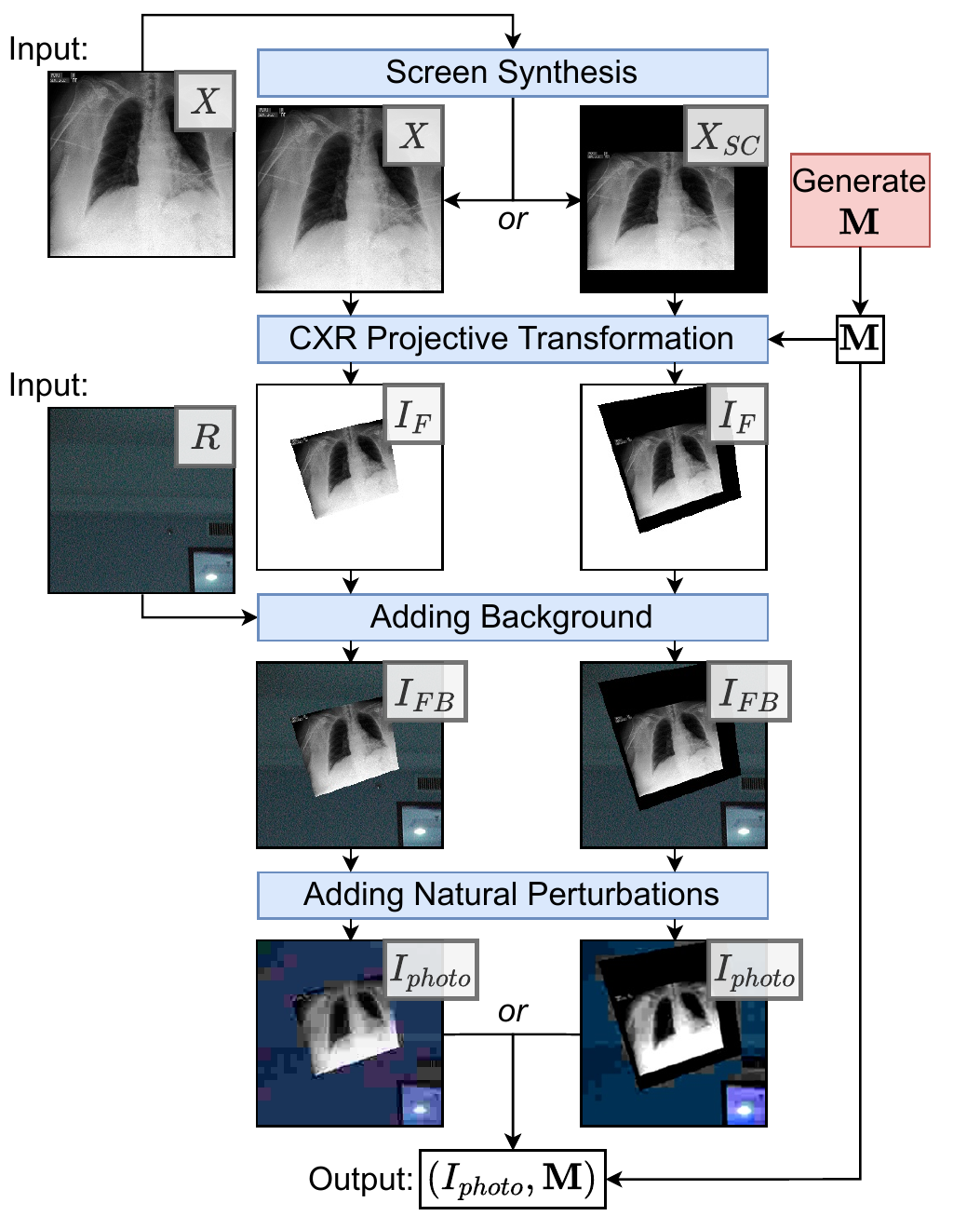}
    \caption{The synthetic data generation framework.}
    \label{fig:framework}
\end{figure}

The generation of one synthetic training sample consists of 4 steps, as shown in Figure \ref{fig:framework}: (1) screen synthesis; (2) CXR projective transformation; (3) adding a background image; and (4) adding natural perturbations. The details of the 4 steps are described in the following subsections. A random CXR $X$ and a random image $R$ are required for generating a sample.

Note that the design of the framework does not exactly follow Equation \ref{eqn:1}, as various natural appearances of CXR photos are also considered such as screen and natural perturbations. Randomness is involved in the framework to increase the diversity of the synthesized CXR photos so that hopefully the trained PTRN can be generalized to photos captured in various scenarios. An ablation study is conducted in Section \ref{sec: Ablation Study on Training Sample Synthesis} to verify the framework. The framework only uses general image processing methods and general augmentation methods. Through experiment, the generation speed attains 102.89 samples/second in a PC with Intel i9-10900K CPU.

\subsubsection{Screen Synthesis}

\begin{figure}
    \centering
    \includegraphics[width=0.5\textwidth]{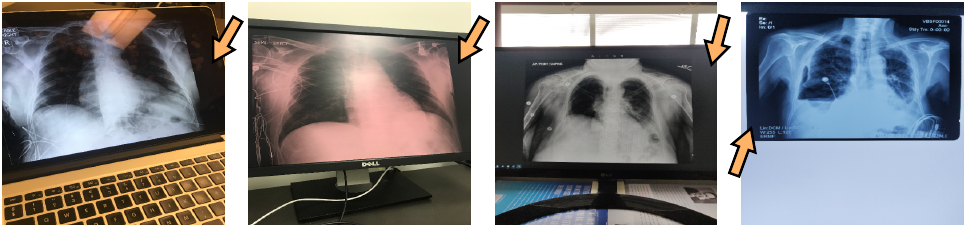}
    \caption{The CXRs in photos are usually surrounded with a dark area.}
    \label{fig:natural_screen}
\end{figure}

\begin{figure}
    \centering
    \includegraphics[width=0.5\textwidth]{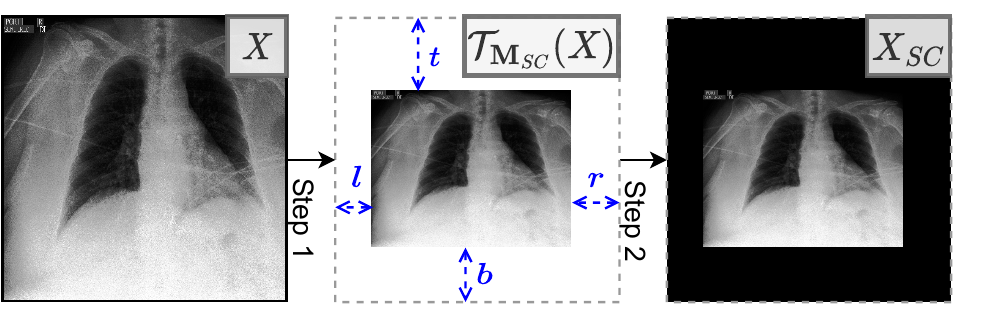}
    \caption{The two sub-steps of screen synthesis.}
    \label{fig:screen_synthesis}
\end{figure}

In a natural CXR photo, the transformed CXR is often surrounded by dark padding, which is usually a monitor screen, as shown in Figure \ref{fig:natural_screen}. As the dark padding looks like a part of the CXR. It may interfere with the prediction of PTRN. This step is to synthesize the screen to simulate such a situation to improve the robustness of PTRN in this situation. Considering that not all natural CXR photos have such a situation, the framework is designed to sometimes synthetic the screen. In experiments, the probability is set to be $0.3$. If a screen is synthesized, the output of this step is a CXR with a synthetic screen $X_{SC}$, otherwise, the output is identical to the input $X$. The process of screen synthesis has two sub-steps, as shown in Figure \ref{fig:screen_synthesis}.

\textbf{Sub-step 1}: $X$ is scaled-down and translated, determined by the randomly generated padding sizes $t,b,l,r$ (see Figure \ref{fig:screen_synthesis}). In the experiments, we set: $t,b,l,r \sim U_{[0,0.6]}$ . Subsequently, the operation of scaling down and translation is implemented by projective transformation. The transform matrix $\mathbf{M}_{SC}$ is calculated by:
\begin{equation}\label{eqn:4}
    \mathbf{M}_{S C}=\left[\begin{array}{ccc}
    1-(l+r) / 2 & 0 & (l-r) / 2 \\
    0 & 1-(t+b) / 2 & (t-b) / 2 \\
    0 & 0 & 1
    \end{array}\right]
\end{equation}
. Finally, $X$ is scaled-down and translated: $\mathcal{T}_{\mathbf{M}_{SC}}(X)$.

\textbf{Sub-step 2}: $\mathcal{T}_{\mathbf{M}_{SC}}(X)$ and a plain dark color image $I_{color}$ are composited to get the output $X_{SC}$:
\begin{equation}
    X_{SC}=\mathcal{T}_{\mathbf{M}_{SC}}(X) \oplus I_{color}
\end{equation}
. In the experiments, the RGB values of the $I_{color}$ is randomly generated by $R, G, B \sim U\left\{0, 19\right\}$.

\subsubsection{CXR Projective Transformation}

This step is to simulate the projective transformation caused by the non-ideal camera position. Firstly, a projective transformation matrix $\mathbf{M}$ is randomly generated. Then, projective transformation with $\mathbf{M}$ is applied to the output of step 1 to get a transformed CXR $I_F$.
The 8 parameters $\theta_i$ of $\mathbf{M}$ must be generated under some regulations since an inappropriate set of parameters results in a nonsensical transformed CXR, e.g., the shape is non-convex \cite{hartley1999theory,solomon2011fundamentals}. To avoid such unexpected situations, we first equivalently represent a projective transformation by a sequential series of five individual actions: scaling, shearing, rotation, perspective warp, and translation. Then, the parameters of these actions are generated within an appropriate range. Specifically, when generating a matrix $\mathbf{M}$, the first step is to generate five 3-by-3 matrices that represent the five individual actions: 
\begin{equation}
    \mathbf{M}_C=\begin{bmatrix}
        C_x & 0 & 0 \\ 0 & C_y & 0 \\ 0 & 0 & 1
    \end{bmatrix}
    \mathbf{M}_S=\begin{bmatrix}
        1 & S_x & 0 \\ S_y & 1 & 0 \\ 0 & 0 & 1
    \end{bmatrix},
    \mathbf{M}_R=\begin{bmatrix}
        \cos \alpha & \sin \alpha & 0 \\ -\sin \alpha & cos \alpha & 0 \\ 0 & 0 & 1
    \end{bmatrix},
\end{equation}
\begin{equation}
    \mathbf{M}_P=\begin{bmatrix}
        1 & 0 & 0 \\ 0 & 1 & 0 \\ P_x & P_y & 1
    \end{bmatrix},
    \mathbf{M}_T=\begin{bmatrix}
        1 & 0 & T_x \\ 0 & 1 & T_y \\ 0 & 0 & 1
    \end{bmatrix},
\end{equation}
where $\mathbf{M}_C,\mathbf{M}_S,\mathbf{M}_R,\mathbf{M}_P,\mathbf{M}_T$ denote the transform matrix of scaling, shearing, rotation, perspective warp, and translation, respectively. The parameters in the five metrices are randomly generated in appropriate ranges. The setup in the experiment is shown in Table \ref{tab:params}.
Then, the generated $\mathbf{M}$ is calculated by:
\begin{equation} \label{eqn:6}
    \mathbf{M}=\mathbf{M}_T \mathbf{M}_P \mathbf{M}_R \mathbf{M}_S \mathbf{M}_C
\end{equation}

\begin{table}
    \centering
    \footnotesize
    \begin{tabular}{|c|c|}
        \hline
        \textbf{Parameters} & \textbf{Generation} \\
        \hline
        Scaling & $C_x,C_y \sim U_{[0.2,0.8]}, |C_x-C_y|\le 0.2$\\
        \hline
        Shearing & $S_x, S_y \sim U_{\left[-0.1, 0.1\right]}$\\
        \hline
        Rotation & $\alpha \sim U_{[-\pi,\pi]}$\\
        \hline
        Perspective Warp & $F_x, F_y \sim N{\left(\mu=0, \sigma^2=0.1^2\right)}$\\
        \hline
        Translation & $T_x, T_y \sim N{\left(\mu=0, \sigma^2=0.25^2\right)}$\\
        \hline
    \end{tabular}
    \caption{Setup in experiment for generating the parameters of the five individual actions. The setup of$|C_x-C_y|\le 0.2$ is to avoid the transformed CXRs be too narrow.}
    \label{tab:params}
\end{table}

The generated $\mathbf{M}$ is then used to transform the output of step 1 to calculate the transformed CXR $I_F$. The output of step 1 has two types: $X$ and $X_{SC}$. In the case of $X$, $I_F$ is calculated by
\begin{equation}
    I_F=\mathcal{T}_\mathbf{M}\left(X\right)
\end{equation}
; In the case of $X_{SC}$, since the CXR region in $X_{SC}$ has been transformed by $\mathbf{M}_{SC}$, the transformation by $\mathbf{M}_{SC}$ must be inversed to ensure that $\mathbf{M}$ can correctly represent the CXR region in $I_F$. The transformed CXR $I_F$ is calculated by
\begin{equation}
    I_F=\mathcal{T}_{\mathbf{M} \mathbf{M}_{SC}^{-1}}\left(X_{SC}\right)
\end{equation}
.

\subsubsection{Adding Background Image}

In practical situations, the background of CXR photos varies. We consider it as a random image $R$. This step is to composite the output $I_F$ of step 2 and a background image. The output $I_{FB}$ is calculated by $I_{FB}=I_F\oplus R$.

\subsubsection{Adding Natural Perturbations}

This step is to simulate the perturbations of natural photos. In the experiments, various data augmentation methods are used to simulate the perturbations, as listed in Table \ref{tab:dataaug}.

\begin{table}
    \centering
    \footnotesize
    \begin{tabular}{|c|c|}
        \hline
        \textbf{Perturbations} & \textbf{Data Augmenration Methods} \\
        \hline
        \multirow{6}{6em}{Illuminations} & Adding pixel values \\
        \cline{2-2}
        & Multiplying pixel values \\
        \cline{2-2}
        & Adding hue and saturation value in HSV\\
        \cline{2-2}
        & Color Enhancement\\
        \cline{2-2}
        & Brightness Enhancement\\
        \cline{2-2}
        & Sharpness enhancement\\
        \hline
        Out-focus & Average blur \\
        \hline
        Image noises& Adding Gaussian noise \\
        \hline
        Image compression& JPEG compression\\
        \hline    
    \end{tabular}
    \caption{Data augmentation methods for adding natural perturbations.}
    \label{tab:dataaug}
\end{table}

\begin{figure}
    \centering
    \includegraphics[width=0.5\textwidth]{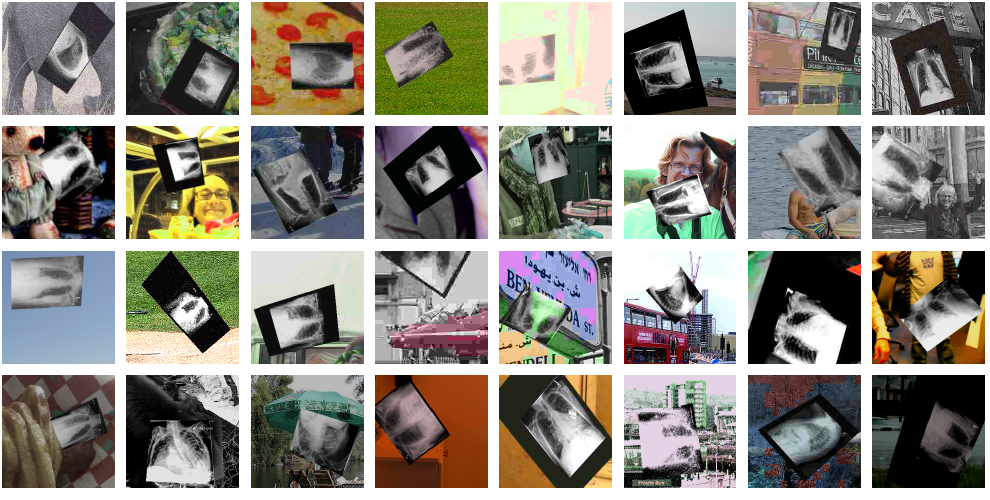}
    \caption{Synthetically generated CXR photos.}
    \label{fig:synthetic_samples}
\end{figure}

In the end, $I_{photo}$ and matrix $\mathbf{M}$ are composited to be a synthetic training sample $(I_{photo},\mathbf{M})$. Figure \ref{fig:synthetic_samples} shows some samples of generated $I_{photo}$.

\section{Experimental Results and Discussion} \label{sec:Experimental Results and Discussion}

All experiments are conducted on a desktop computer with hardware Intel i9-10900KF CPU, 128GB RAM, and RTX 3090 24G GPU. The deep learning platform is TensorFlow 2 \cite{tensorflow2015-whitepaper} on Python 3, installed in a Linux Mint 20.1 OS. The data augmentation methods are implemented using imgaug library \cite{imgaug}.

\subsection{Datasets}

\begin{figure}
    \centering
    \includegraphics[width=0.5\textwidth]{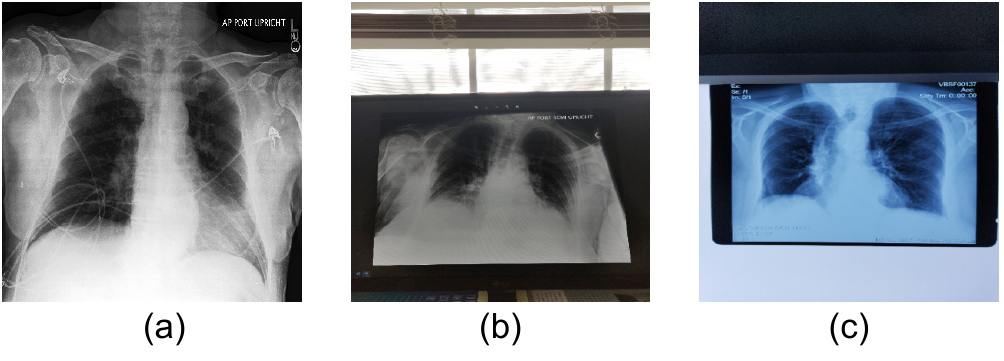}
    \caption{(a) CheXpert validation set. (b) CheXphoto-Monitor validation set. (c) CheXphoto-Film validation set.}
    \label{fig:datasets}
\end{figure}

\subsubsection{CheXpert}

CheXpert \cite{irvin2019chexpert} is a public digital CXR dataset for competition. It consists of 224,316 digital CXRs from 65,240 patients. Each digital CXR is labeled into at least one of 14 pathologies: No Finding, Enlarged Cardiomediastinum, Cardiomegaly, Lung Opacity, Lung Lesion, Edema, Consolidation, Pneumonia, Atelectasis, Pneumothorax, Pleural Effusion, Pleural Other, Fracture, and Support Devices. The validation set consists of 234 digital CXRs from 200 studies (Figure \ref{fig:datasets}(a)).

\subsubsection{CheXphoto}

CheXphoto \cite{phillips2020chexphoto} is a public smartphone-captured CXR photos dataset for competition. It consists of 10,507 CXR photos from 3,000 patients, which are sampled from digital CXRs in CheXpert. The labels are also inherited from CheXpert. The validation set consists of synthesis photos and natural photos. Only natural photos are used for evaluation in this paper. The validation set has two types of natural photos: (1) A total of 234 CXR monitor photos (Figure \ref{fig:datasets}(b)). They are the photo version of the CheXpert validation set. Each photo is produced by using a smartphone camera to capture digital CXRs displayed on a monitor. (2) A total of 250 CXR film photos (Figure \ref{fig:datasets}(c)). Each photo is produced by using a smartphone camera to capture CXR film. In Section \ref{sec:CheXphoto Competition}, PTRN is used for the CheXphoto competition. In the competition, the private test CXR photos were split into two private test sets by type (monitor/film). Therefore, we split the CheXphoto validation set into two sets by the type (monitor/film) for experiments, referred to as the CheXphoto-Monitor validation set and the CheXphoto-Film validation set, respectively. Note that the CheXphoto-Monitor validation set is the photo version of the CheXpert validation set (digital CXR), we use these two sets to demonstrate the effectiveness of PTRN in Section \ref{sec: Classification Performance on Automatically Rectified CXR Photos}.

\subsection{Implementation Details} \label{sec:Implementation Details}
For PTRN, an ImageNet-pretrained DenseNet-201 \cite{huang2017densely} is used as the CNN backbone. Adam optimizer \cite{kingma2014adam} with a learning rate of $1\times 10^{-5}$ and parameters $(\beta_1=0.9,\beta_2=0.999,\epsilon=1\times 10^{-7})$ is used to update the weights. The batch size is set to 32. The input image is a $224\times224$ pixel RGB image. Pixel values are linearly rescaled from $[0,255]$ to $[0,1]$. Among these hyperparameters, only the learning rate was simply tuned while others remain default. The trained PTRN has sufficient rectification performance already (see Section \ref{sec: Ablation Study on Training Sample Synthesis}), which demonstrates the easy training of PTRN.

For synthetic training data generation, we use the 224,316 digital CXRs from the CheXpert training set as the source of the CXR images, and the 41K images from the Microsoft COCO 2017 test set \cite{lin2014microsoft} as the source of the random images. When generating a sample, a CXR image and a random image are randomly picked from the sources. Benefitting from the fast generation speed of synthetic data, we dynamically generate the synthetic training data during training PTRN. Therefore, each training sample is used only once to avoid overfitting.

The performance of PTRN is validated on the CheXphoto-Monitor validation set per 100 weights updates. The checkpoint with the highest IoU is picked and furtherly tested on the CheXphoto-Film validation set. The evaluation of PTRN requires the 4 vertices of the ground truth region for each sample. Since the CheXphoto dataset did not provide these annotations, we manually annotated the 4 vertices of the CXR photos.

\subsection{CXR Photos Classification Pipeline} \label{sec: CXR Photos Classification Pipeline}

\begin{figure}
    \centering
    \includegraphics[width=0.5\textwidth]{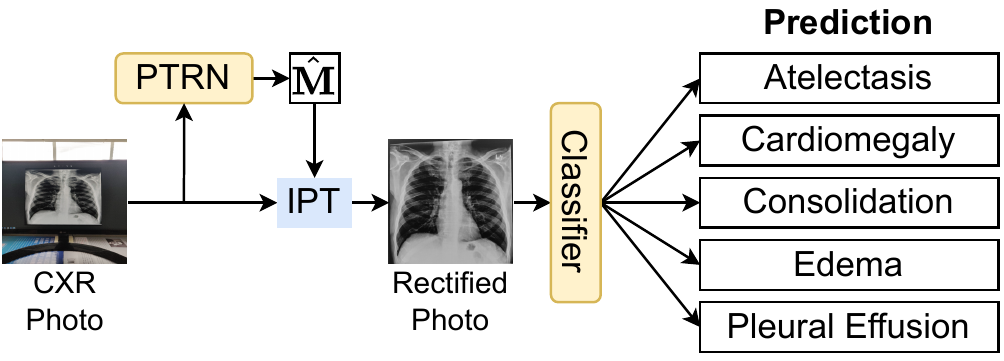}
    \caption{CXR photos classification pipeline.}
    \label{fig:pipeline}
\end{figure}

The pipeline to perform CXR photos classification (Figure \ref{fig:pipeline}) consists of three steps: (1) PTRN predicts the projective transformation matrix $\hat{\mathbf{M}}$ of a CXR photo; (2) the photo is rectified using the predicted  $\hat{\mathbf{M}}$; (3) A classifier trained on high-quality digital CXRs evaluates the rectified CXR photo. This pipeline is used in the experiments in Sections \ref{sec:CheXphoto Competition} and \ref{sec: Classification Performance on Automatically Rectified CXR Photos}.

\subsection{CheXphoto Competition} \label{sec:CheXphoto Competition}

CheXphoto is a competition for smartphone-captured CXR photos classification hosted by Stanford and VinBrain \cite{phillips2020chexphoto}. A submitted model is tested to perform multi-label classification on the two private test sets, respectively:
\begin{enumerate}
    \item CXR film photos. A total of 250 CXR films are captured as photos by a smartphone camera. The ranking of the leaderboard is sorted in terms of AUC on this private test set. We refer to this set as CheXphoto-Film private test set.
    \item CXR monitor photos. A total of 668 digital CXRs from 500 studies are displayed on a monitor and captured as photos by an iPhone 8. We refer to this set as CheXphoto-Monitor private test set.
\end{enumerate}
The performance of a model is measured by calculating the mean AUC-ROC score on 5 selected pathologies: Atelectasis, Cardiomegaly, Consolidation, Edema, and Pleural Effusion.

We build a classifier for this competition with the pipeline in Section \ref{sec: CXR Photos Classification Pipeline}. The classifier consists of 5 weighted average ensembles. Each ensemble outputs the predicted probability of a pathology. Each ensemble is composed of 4 to 6 trained single CNN classifiers for binary classification. A single classifier is a CNN model with an input of $224\times224$ pixel grayscale image. This setup does not like other previous work, a single classifier only predicts one pathology instead of all 5 pathologies. It is because the numbers of training samples across the 5 pathologies are quite unbalanced. While training a multi-label classifier, the overfitting timings of the 5 pathologies are different. Therefore, we employ binary classifiers to obtain optimal performance for each pathology. The single classifiers are trained on the CheXpert training set with different configurations to improve the performance of the constructed ensembles \cite{zhou2012ensemble,hansen1990neural}. The different configurations include different CNN models, learning rates, batch sizes, configurations of label smoothing regularization from \cite{pham2021interpreting} for handling uncertainty labels, and data augmentation methods such as adjusting brightness, contrast, and adding Gaussian noise. Cross entropy loss is used to calculate the loss. The validation dataset is the CheXphoto-Film validation set in which the projective transformation is perfectly rectified by manual operation.

The results are reported in Table \ref{tab:chexphoto}. our pipeline outperforms all published work \cite{le2020interpretation,chong2021gan} in the CheXphoto dataset. Moreover, our pipeline achieves \textbf{first place} on the CheXphoto competition leaderboard, as shown in Table \ref{tab:competition}. The first, second, and third places are all our work based on PTRN. We compare our pipeline with the fourth place, our approach yields a huge performance gap on CXR film photos (0.850 vs. 0.762, 0.088 higher in AUC).

\begin{table}
    \centering
    \footnotesize
    \begin{tabular}{c||cc|cc}
        \hline
        \multirow{2}{4em}{Method} & \multicolumn{2}{c|}{CXR Film Photos} & \multicolumn{2}{c}{CXR Monitor Photos} \\
        \cline{2-5}
        & Validation & Test & Validation & Test \\
        \hline

        YOLOv3 \cite{le2020interpretation} & - & - & 0.684 & - \\
        GAN-STAM \cite{chong2021gan} & - & - & 0.865 & - \\
        \textbf{PTRN (Ours)} & \textbf{0.868} & \textbf{0.850} & \textbf{0.885} & \textbf{0.891} \\
        \hline
    \end{tabular}
    \caption{Quantitative results in AUC of PTRN on the CheXphoto validation sets and the CheXphoto competition private test sets.}
    \label{tab:chexphoto}
\end{table}

\begin{table}
    \centering
    \footnotesize
    \begin{tabular}{cccc}
        \hline
        Rank & Model & CXR Film Photos & CXR Monitor Photos \\
        \hline
        \textbf{1} & \textbf{LBC-v2} & \textbf{0.850} & \textbf{0.89} \\
        \textbf{2} & \textbf{LBC-v0} & \textbf{0.820} & \textbf{0.89} \\
        \textbf{3} & \textbf{Stellarium-CheXpert-Local} & \textbf{0.802} & \textbf{0.88} \\
        4 & MVD121 & 0.762 & 0.83 \\
        5 & MVD121-320 & 0.758 & 0.84 \\
        \hline
    \end{tabular}
    \caption{The top 5 of the leaderboards of the CheXphoto competition (22 Nov 2022). The top 3 spots are all our PTRN.}
    \label{tab:competition}
\end{table}

\subsection{Classification Performance on Automatically Rectified CXR Photos} \label{sec: Classification Performance on Automatically Rectified CXR Photos}

To verify the classification performance improvement in the CheXphoto competition is mostly contributed by PTRN instead of the ensembling, we compare the classification performance of a single CNN classifier with/without PTRN on CXR photos. The single CNN classifier is an Xception \cite{chollet2017xception} trained on the CheXpert dataset. It achieves AUC 0.889/0.887 on the CheXpert validation/test set, respectively. The classification performance is close to the single model of \#2 \cite{pham2021interpreting} in the CheXpert competition (validation AUC 0.894). The CNN has been uploaded to the CheXphoto competition. In the leaderboard, the name of the CNN without PTRN is \textit{Stellarium}, and the name of the CNN with PTRN is \textit{Stellarium-CheXpert-local}. The results are shown in Table \ref{tab:ptrn}.

\begin{table}
    \centering
    \footnotesize
    \begin{tabular}{c||cc|cc|cc}
        \hline
        \multirow{2}{4em}{Method} & \multicolumn{2}{c|}{CheXpert} & \multicolumn{2}{c|}{CheXphoto-Monitor} & \multicolumn{2}{c}{CheXphoto-Film} \\
        \cline{2-7}
        & Validation & Test & Validation & Test & Validation & Test\\
        \hline
        CNN & 0.889 & 0.887 & 0.821 & 0.710 & 0.722 & 0.599 \\
        \textbf{PTRN+CNN} & \textbf{0.893} & \textbf{0.896} & \textbf{0.893} & \textbf{0.880} & \textbf{0.791} & \textbf{0.802} \\
        \hline
    \end{tabular}
    \caption{Quantitative results in AUC of PTRN on the CheXphoto validation sets and the CheXphoto competition private test sets.}
    \label{tab:ptrn}
\end{table}

The first row reports the performance of the CNN without PTRN, huge performance drops of this CNN are observed on the CXR monitor photos and CXR film photos, since the photos experienced projective transformation with extra noises. In the second row, PTRN is used to rectify the projective transformation of CXR photos. The classification pipeline follows the one in Section \ref{sec: CXR Photos Classification Pipeline}.

For the results on the CXR monitor photos, it achieves AUC 0.893/0.880 on the validation/test set, respectively, which is far superior to the AUC scores before rectification (AUC 0.821/0.710). The data in the CheXphoto-Monitor validation set is the photos version of the digital CXRs in the CheXpert validation set. The performance on the CheXphoto-Monitor validation set (AUC 0.893) is the same as the performance on the CheXpert validation set (AUC 0.893). It indicates that PTRN can maintain the classification performance on CXR photos to the same level as on digital CXRs. It implies that PTRN is sufficient to eliminate all the negative impacts of projective transformation to classification performance.

In CXR film photos, huge performance improvements are observed after using PTRN (AUC 0.802/0.599, 0.203 improvement on the CheXphoto-Film private test set), which verifies the effectiveness of the PTRN on both types of photos.

Besides, we furtherly test the pipeline in the CheXpert dataset. Surprisingly, minor improvements are also observed. A further investigation is needed to verify the performance impact of PTRN on digital CXRs.

\subsection{Ablation Study on Training Sample Synthesis} \label{sec: Ablation Study on Training Sample Synthesis}

The synthetic data framework consists of 4 steps. However, based on the formulation of Equation \ref{eqn:1}, a synthetic CXR photo can be composited by only a transformed CXR and a background. It means in the framework, only step 2 (projective transformation) and 3 (adding background) are necessary. Step 1 (screen synthesis) and step 4 (adding natural perturbations) are additional to simulate the appearances of natural CXR photos. To study the impact of steps 1 and 4 to the rectification performance, we conducted an ablation study by removing step 1 or/and 4.

\begin{table}
    \centering
    \footnotesize
    \label{tab:framework}
    \begin{tabular}{cccc||cc}
        \hline
        \multicolumn{4}{c||}{Step} & CheXphoto-Monitor Validation & CheXphoto-Film Validation \\
        \cline{1-4}
        1 & 2 & 3 & 4 & \textit{(for validation)} & \textit{(for test)}\\
        \hline
        \textbf{\checkmark} & \textbf{\checkmark} & \textbf{\checkmark} & \textbf{\checkmark} & \textbf{0.942 (0.938, 0.946)} & \textbf{0.892 (0.887, 0.898)} \\
        & \checkmark & \checkmark & \checkmark & 0.911 (0.905, 0.916) & 0.816 (0.808, 0.824) \\
        \checkmark & \checkmark & \checkmark & & 0.919 (0.914, 0.924) & 0.800 (0.790, 0.809) \\
        & \checkmark & \checkmark & & 0.811 (0.875, 0.886) & 0.815 (0.807, 0.824) \\
        \hline

    \end{tabular}
    \caption{Ablation study on the framework for generation of synthetic data (In mean IoU with 95\% C.I.)}
\end{table}

The results are shown in Table 5. Mean IoU (mIoU) with 95\% confidence interval (C.I.) is reported. In the first row, the mIoU of PTRN that is trained with all four steps is reported. It achieves mIoU 0.942/0.892 on the validation/test set respectively. After removing the step(s), the mIoU on the validation set are dropped by approximately 0.02-0.07, and the mIoU on the test set is dropped much larger (approximately 0.07-0.10). It demonstrates the necessity and importance of step 1 and 4. It also verifies that the framework for the generation of synthetic CXR photos has a sufficient variance of the key variables controlling the transformed CXRs.

\subsection{Qualitative Evaluation}

\begin{figure}
    \centering
    \includegraphics[width=0.5\textwidth]{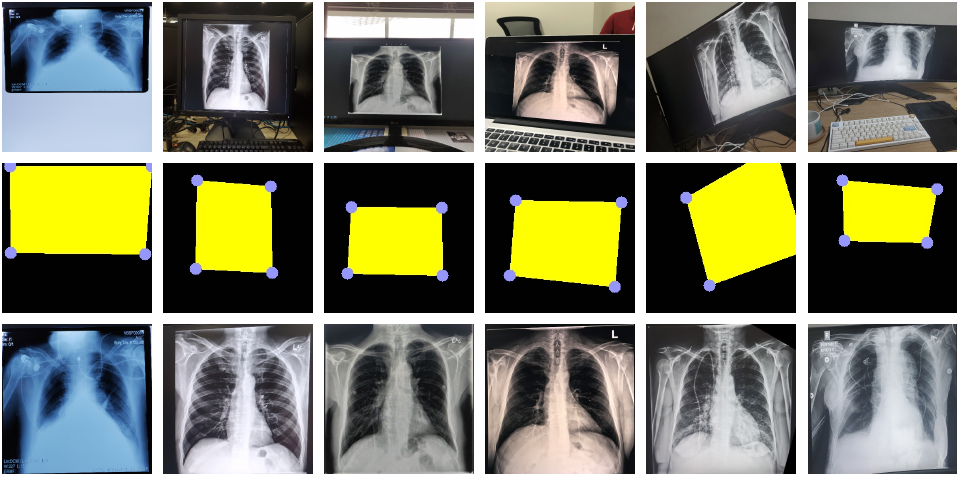}
    \caption{Top row: CXR photos. Mid row: Prediction of PTRN. Bot row: CXR photos that are rectified by using the prediction.}
    \label{fig:examples}
\end{figure}

In Figure \ref{fig:examples}, we demonstrate using trained PTRN to automatically rectify six CXR photos captured in different scenarios. All six CXR photos are properly rectified. These results demonstrate that PTRN is robust to CXR photos captured in different scenarios.

\section{Conclusion} \label{sec:Conclusion}

In this paper, we present a deep learning-based Projective Transformation Rectification Network (PTRN) that is trained with synthetic data for rectifying the projective transformation of CXR photos. We also propose a framework to generate synthetic data. Our pipeline achieves first place on the CheXphoto competition leaderboard, which has a significant improvement over \cite{le2020interpretation} and \cite{chong2021gan}. This result verifies the performance of our PTRN and the framework for the generation of synthetic data. Additionally, the design of PTRN and the framework for the generation of synthetic data can be applied to other image recognition fields that encounter similar image distortion caused by the imperfect camera position.

This work has certain limitations. For example, in Section \ref{sec:Implementation Details}, the hyperparameters were not particularly tuned in the training of PTRN since it already produced satisfying rectification performance. The performance could be further improved by tuning hyperparameters or choosing other CNNs as the backbone. Moreover, the CXR photos classification pipeline consists of two CNN models, which requires a higher computation cost. The above limitations may lead to possible directions to extend or improve this work.

\section*{Acknowledgment}

This work is supported by Macao Polytechnic University under grant number RP/ESCA-01/2021.





\bibliographystyle{elsarticle-num}
\bibliography{bib}




\end{document}